\documentclass[a4paper]{spie}
\usepackage[latin1]{inputenc}
\usepackage{amsmath}
\usepackage{amsfonts}
\usepackage{amssymb}
\usepackage[]{graphicx}
\usepackage{color}%,ulem}

\title{Efficient single image non-uniformity correction algorithm}
\author{Y. Tendero \supit{a}, J. Gilles\supit{a,c}, S. Landeau\supit{c}, and J.M. Morel \supit{a} \skiplinehalf
            \supit{a} Centre de Math\'ematiques et de Leurs Applications (CMLA)\\
	  \'Ecole Normale Sup\'erieure de Cachan - 61 av du Pdt Wilson 94235 Cachan Cedex France.\\
           \supit{c} D\'el\'egation G\'en\'erale pour l'Armement, Centre d'Expertise Parisien - 7 Rue des Mathurins, 92221 Bagneux.          
           }
\authorinfo{Further author information: (Send correspondence to Y. Tendero)\\E-mail: tendero@cmla.ens-cachan.fr, Phone: +33 1 47 40 59 48\\
}
\begin{document}
\maketitle
\begin{abstract}
This paper introduces a new way to correct the non-uniformity (NU) in uncooled infrared-type images. The main defect of these uncooled images is the lack  of a column (resp. line) time-dependent cross-calibration, resulting in a strong column (resp. line) and time dependent noise.  This problem can be considered as a 1D flicker of the columns inside each frame. Thus, classic movie deflickering algorithms can be adapted, to equalize the columns (resp. the lines). The proposed method therefore applies to the series formed by the columns of an infrared image a movie deflickering algorithm. The obtained  single image method works on static images, and therefore requires  no registration, no camera motion compensation, and no closed aperture sensor equalization. Thus, the method has only one camera dependent parameter, and is landscape independent. This simple method will be  compared to a state of the art total variation single image correction on raw real and simulated images. The method is real time, requiring only two operations per pixel. It involves no test-pattern  calibration and produces no ``ghost artifacts''.
\end{abstract}

\section{Introduction}
\label{sec:intro}  % \label{} allows reference to this section

Infrared (IR) imaging has proved to be a very efficient tool in a wide range of industry, medical, and military applications. IR cameras are used to measure temperatures, IR signatures, detection, etc.
However, the performance of the imaging system is strongly affected by a random spatial response of each pixel sensor. Under the same illumination the readout of each sensor is different. This is due to mismatches in the fabrication process, among other issues \cite{book_electro_optical}. Furthermore for uncooled cameras the problem is even worse because the  sensor response non-uniformity is not stationary and slowly drifts in time. For this kind of camera a  periodic update of the non-uniformity-correction (NUC) is required.\\
A good non-uniformity-correction is a key success factor for any post processing such as pattern recognition, image registration, etc. To get the rid of the non-uniformity, two main kinds of methods have been developed:
\begin{itemize}
\item Calibration based techniques consist in an equalization of the response to an uniform black body source radiation. They are not convenient for real time applications, since they force to interrupt the image flow. (This calibration is usually automatic, a shutter closing in front of the lens periodically).
\item Scene based techniques, involving motion compensation or temporal accumulation. Such methods are complex and require certain observation conditions.
\end{itemize}
The perturbation model is $$z_t(X)= f_{(X,t)}(u0_t(X))+ \eta_t(X)$$
where ($X$ is the position and $t$ is the time for the following) $z_t(X)$ is the observed value, $u0_t(X)$ is the ideal landscape, $f_{(X,t)}$ is the (unknown) transfer function of the sensor, and $\eta_t(X)$ is a random sensor Poisson noise. A non-uniformity correction algorithm aims at discovering $f_{(X,t)}$ or $u0_t(X)$ for each $X$ and $t$.
In this paper we propose a single frame based algorithm and show that motion compensation or accumulation algorithms are not necessary to achieve a good image quality. However, the proposed method can be viewed as a first step fostering the success of more sophisticated motion based correction algorithms. These are slow while the proposed algorithm is real time, and the obtained quality after a single frame correction might be sufficient for many uses.  \\
The paper is organized as follows. Section \ref{sec:related} presents  related works. The new algorithm is described in section \ref{sec:our}. Experiments on simulated, real, cooled and uncooled images are described in section \ref{sec:experiments}. Section \ref{sec:conclusion} contains a discussion. Possible improvements are envisaged in section \ref{sec:improvement}.

\section{Anterior work}
\label{sec:related}
Numerous algorithms have been reported in the literature to remove the fixed-pattern-noise caused by the lack of a cross-column sensor equalization. Some algorithms estimate the sensor parameter and others attempt at recovering the true landscape. Most of them use a simplified (linear) model for the transfer function of the pixel sensor: $$z_t(X)=u0_t(X)g_t(X) + b_t(X) + \eta_t(X),$$
 where where ($X$ is the position and $t$ is the time for the following) $z_t(X)$ is the observed value, $u0_t(X)$ is the landscape, $g_t(X)$ and $b_t(X)$ are the gains/ offsets (in place of $f_{(X,t)}$) and $\eta_t(X)$ is the random noise. (Nevertheless, the true transfer function is non linear.) 
These algorithms process a sequence of images $(z_t)_{t \in {1,...,N}}$, not a single frame.
The proposed algorithm uses no registration, hence we will focus on single frame  algorithms. There are methods \cite{bb7630} suggesting to equalize the mean and standard deviation ($stddev$) of each pixel sensor by a linear transform. The key idea is 
 
 \medskip [$\cal{H}$:] If all pixel sensors have seen the same landscape, they should have (at least) the same mean and same standard deviation, namely

$$\left\{ \begin{array}{rr}
   mean \underset{t \in \left( 1,...N \right)}{} \left( z_t(X)\right) &= C_m ~ \forall ~X \\
   stddev \underset{t \in \left(1,...N \right)}{} \left( z_t(X)\right) &= C_{std} ~ \forall ~X.\\
   \end{array}
\right.
$$ So the authors suggest to adjust the sensor readout using a linear transform to obtain the equalities above. But this is only possible if there is a long camera  sequence with enough motion where each sensor sweeps many different parts of the scene. \\
A variant \cite{PezoaTCR04} adjusts the minimum and the maximum of the readout values, assuming the time histograms observed in each sensor to become equal over a long enough time sequence:

$$\left\{ \begin{array}{rr}
   min\underset{t \in \left( 1,...N \right)}{} \left( z_t(X)\right) &= C_1~\forall~X \\
   max\underset{t \in \left( 1,...N \right)}{} \left( z_t(X)\right) &= C_2 ~\forall ~X. \\
   \end{array}
\right.
$$

This last method is called Constant Range \cite{Torres02scene-basednon-uniformity}. As pointed out by several authors \cite{Harris93minimizingthe} the length  $N$ of the sequence is a crucial factor of success here. Two problems may arise:
\begin{itemize}
\item If $N$ is too small and the estimation is wrong because all sensors have not seen the same landscape;
\item If $N$ is too large  and because of the  approximation bias and time drift of the sensor behavior, the previous images may appear as ghosts in the last ones. This undesirable effect is known a ``ghost artifact''.
\end{itemize}
  There is a way to avoid the ghost artifacts \cite{Harris93minimizingthe}, which consists in a reset of the estimation when the scene changes too much. Their \cite{Harris93minimizingthe} paper uses a simple threshold to perform scene change detection. But again, all this requires a long exposition time with a varying scene or a serious camera motion.  \\

There are several implementations for these two major algorithms. A recursive filter \cite{bb7630} estimates the parameters of the linear function which approximates the S-shaped transfer function of the sensor, or a Kalman filter \cite{Torres:03} is preferred.
 Other authors \cite{scribner91,nuc_torres_2003} propose a neural network based algorithm, which requires a serious computational power and is definitely not real time.
 The registration based algorithms \cite{tzimopoulu98} consider often only translations (but homographies should be used instead, at least on a static scene). Creating a panorama has been proposed \cite{hardie00} to obtain a ground truth, and to use it as a calibration pattern. However, as pointed out \cite{DBLP:conf/icip/ZhaoZ06}, in presence of the structured fixed-pattern-noise occurring in  most IR cameras, the panorama won't lead to a good result.

\section{ Midway infrared correction}
\label{sec:our}
\subsection{The midway histogram equalization method}
\begin{figure}[!h]
\begin{center}
\includegraphics[width=9cm]{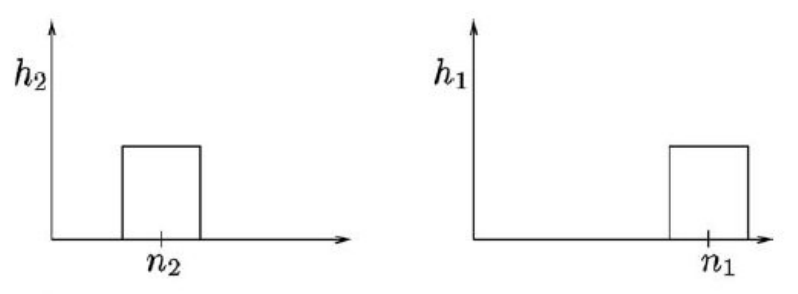}
\includegraphics[width=4cm]{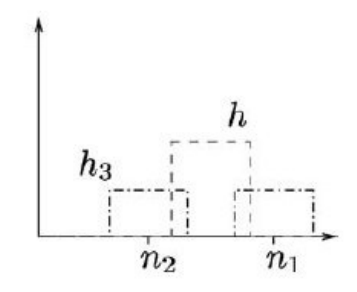}
\end{center}
\caption{\label{mode}Two histograms $h_1$, $h_2$ (left side) and the corresponding midway histogram $h$  (on the right), compared to the direct histogram average, which would create two modes and is therefore wrong.}
\end{figure}

The midway algorithm was designed initially to correct for gain differences between cameras \cite{delon2004midway}. It permits to compare two images taken with different cameras more easily after their histograms have been equalized. This algorithm was later extended to flicker correction \cite{bb77071}.\\
 Consider two cumulative histograms $H_1$, $H_2$.  The midway cumulative histogram of the corrected image is simply  $$Hmid^{-1}=\frac{H_1^{-1}+H_2^{-1}}{2},$$ and this average can be extended to an arbitrary number of images. Once the midway histogram is computed, a monotone contrast change is applied to image to specify $H$ as its histogram. Thus, all images get the midway histogram, which is the best compromise between all histograms (see Fig \ref{mode}).

\subsection{The idea}
 Since many IR correction algorithms actually propose to equalize the temporal histograms of each pixel sensor, the midway is quite adapted to get a better result than a simple equalization. Yet, we propose a still much simpler strategy.
Equalization can be based on the fact that single columns (or lines, depending of the readout system) carry enough information by themselves for an equalization.
The images being continuous, the difference between two adjacent columns is statistically small, implying that two neighboring histograms are nearly  equal. This hypothesis here is similar to the temporal one [$\cal{H}$] but is better suited to the decision to carry the equalization inside the image itself.
So the proposition is to transport  the histogram of each column (or line) to the midway of histograms of neighboring columns (resp. lines). In presence of strong fixed-pattern-noise (FPN) it will be useful to perform this {\it sliding midway method} over a little more than two columns, because the FPN is not independent in general.\\

Assume in the sequel that the equalization is performed with columns.
The proposed algorithm proceeds as follows.

\paragraph{Midway Infrared Equalization (MIRE)}
\begin{itemize}
\item Compute the cumulative histogram $H_i$ of each column $c_i$;
\item For each column $c_i$ compute a local midway histogram  $\tilde H_{mid}(i)^{-1}:=\sum\underset{j \in (-N,...,N)} \Phi(j) H_{i+j}^{-1} $ using Gaussian weights  $\Phi=\Phi_\sigma$ with std-dev $\sigma$ average;
\item Specify the histogram of the column $c_i$ onto this midway histogram $\tilde H_{mid}(i)$.
\end{itemize}

The choice of the standard deviation $\sigma$ of the Gaussian depends only on the camera, and not on the landscape. Thus, it can be fixed once and for ever for each camera. Since we work on images separately the method is not affected by motions or changes of scene, which  completely avoids "ghost artifacts" and any problem caused by the calibration parameters drifting over time. A good $\sigma$ is simply obtained by
\begin{itemize}
\item Trying with a small parameter;
\item Increasing it till a good visual image quality is reached.
\end{itemize}
Yet an automatic method for estimating $\sigma$ and obtaining a parameterless methods is as follows.

\begin{figure}[!h]
\begin{center}
\includegraphics[width=5.5cm]{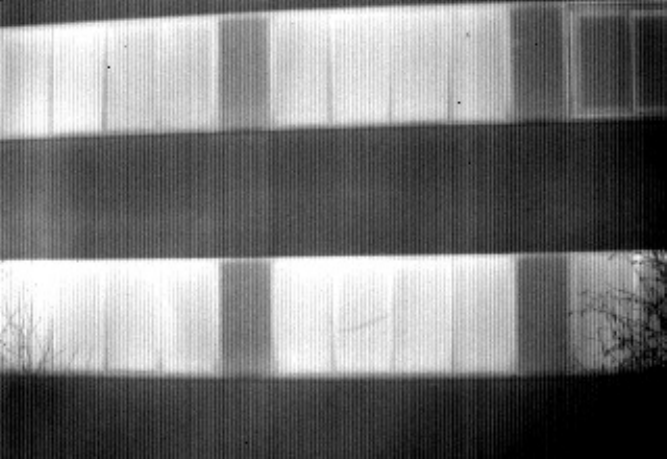}
\includegraphics[width=5.5cm]{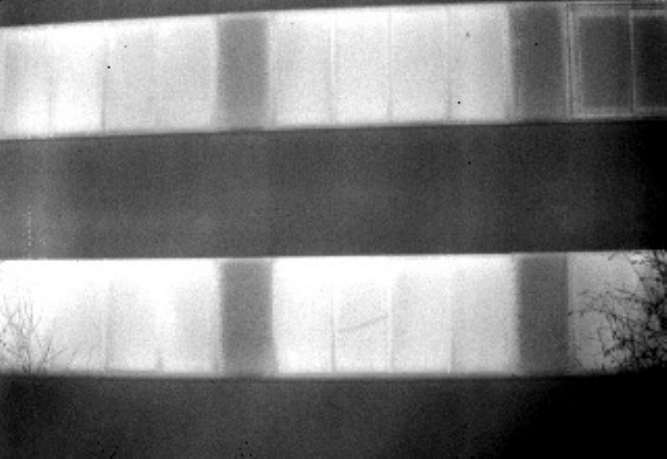}
\includegraphics[width=5.5cm]{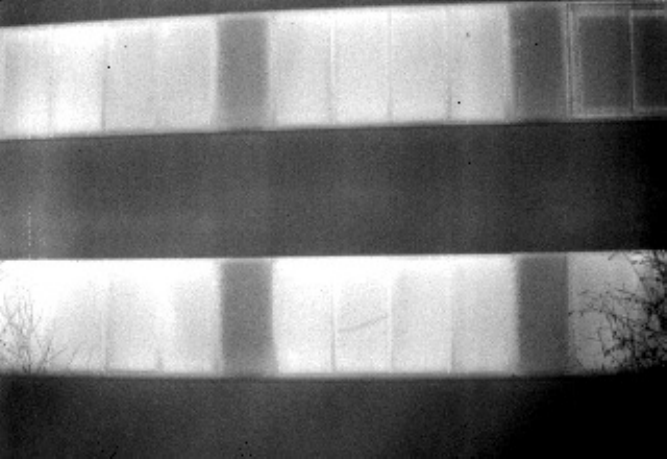}\\
\includegraphics[width=5.5cm]{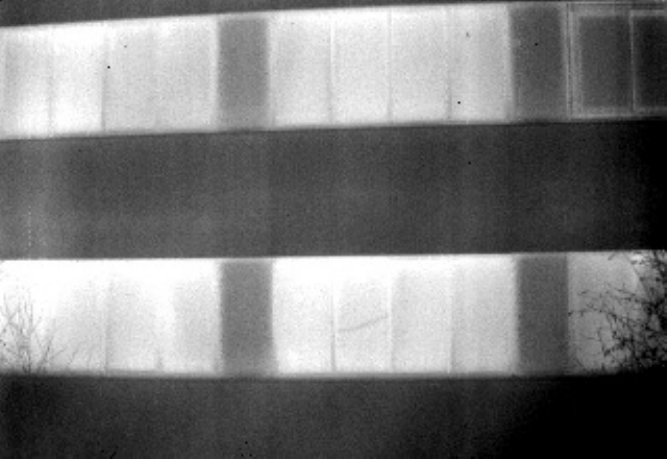}
\includegraphics[width=5.5cm]{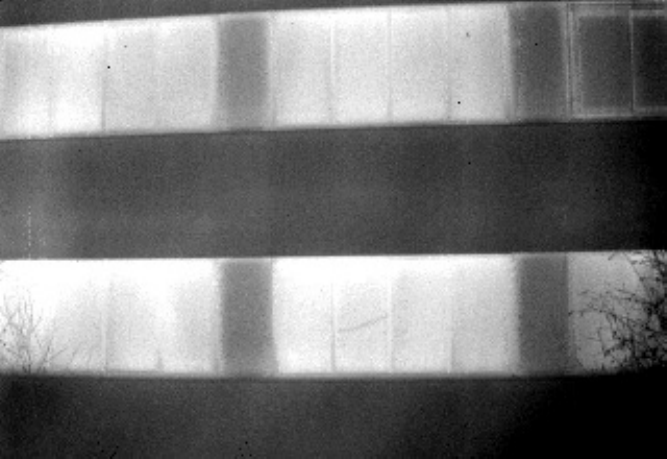}
\includegraphics[width=5.5cm]{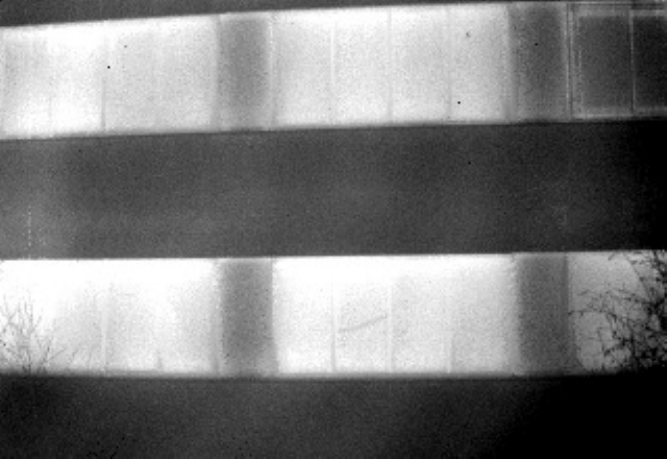}\\
\includegraphics[width=5.5cm]{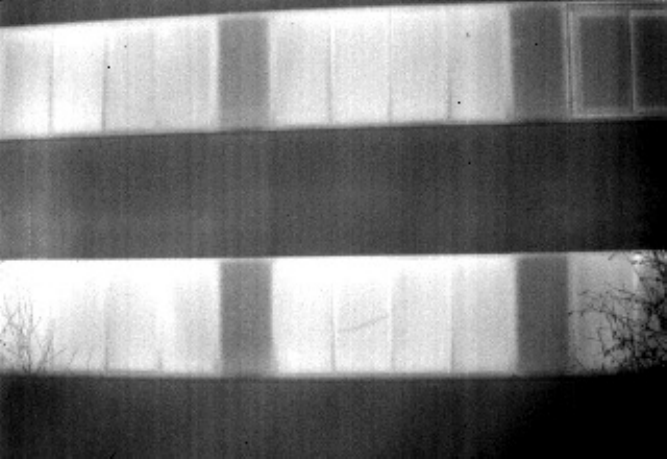}
\includegraphics[width=5.5cm]{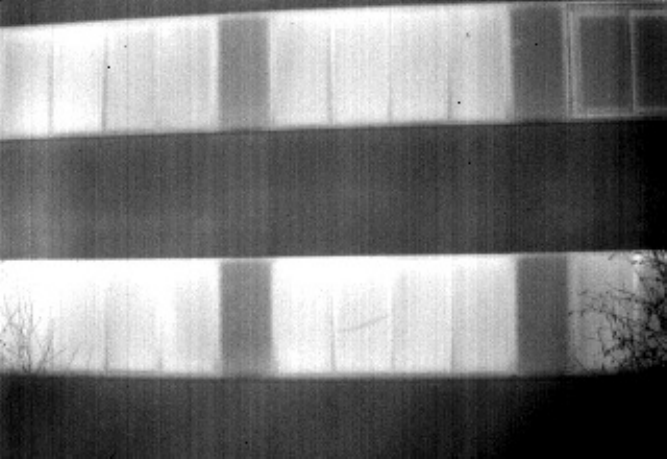}
\includegraphics[width=5.5cm]{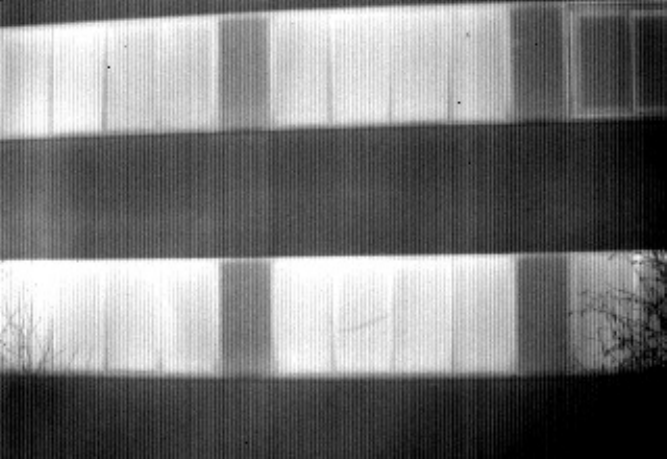}

\caption{\label{sim3}First :RAW (uncooled camera), then from left to right: increasing the $\sigma$ parameter. TV norms : 8.07, 7.44, 7.38, 7.39, 7.36, 7.59, 7.62, 7.85, 8.07.}
\end{center}
\end{figure}

\textbf{Automatically fitting the perfect parameter\\}
The non-uniformity leads to an increased total-variation norm. Hence the smoothest image is also the one with little or no non-uniformity at all. So the simplest way to find the good parameter automatically is :\\
$$\sigma^*=argmin_\sigma||I_\sigma||_{TV} \footnote{see \ref{def:TV} for $||.||_{TV}$ definition.},$$
where $I_\sigma$ is the image processed by MIRE with the parameter $\sigma$.
The optimization could be done by a dichotomy on $\sigma$. See Fig \ref{sim3} for an illustration of this.

\noindent{\bf Theorem 1.}
If  $h_i$ $i \in {1,...,N}$ are $N$ histograms of the same landscape seen by $N$ different columns of the sensor, and $ H_{mid}=\sum_{j=1}^{N} \frac{H_{j}^{-1}} {N} $ then :
$$ ||h_{mid}-h_{true}||_2 \leq max\underset{i \in (1,...,N)}(||h_i-h_{true}||_2)$$
Moreover if the $h_i ~ \forall i \in (1,...,N)$ from the $N$ columns of the sensor are \textit{i.i.d. and centered} on $h_{true}$ then $$||hmid-h_{true}||_2\underset{N \to \infty }{\rightarrow}0$$\\

\subsection{Implementation}
The implementation is easy and was done with \textit{Matlab}. To avoid border effects we used a reflection of the image across borders. The computation times are shown for several image sizes. An on-line demo will  be shortly available at www.ipol.im. \\\\
Times are shown in seconds on a core duo T7250 running Ubuntu and \textit{Matlab}. We used \textit{Timeit} (written by S. Eddins) to avoid time variation of the multitasking OS.\\
\begin{figure}
\begin{center}
\begin{tabular}{|l|c|r|}
  \hline
  Image size & 512*512 & 320*220 \\
  \hline
Seconds & 2.8 & 1.2 \\
  \hline

\end{tabular}
\caption{Computation time for various sizes and quantifications in seconds (using Matlab). This time could be made to real with any standard processor.}
\end{center}
\end{figure}
Of course a temporal extension of the algorithm to avoid temporal flicker is possible, using a temporal midway \cite{bb77071}.

\subsection{Quality analysis}
Our first criterion is the visual image quality. In the simulated cases the results will be evaluated  by the RMSE,
$$ RMSE(I,\tilde I) =  \sqrt{\dfrac{\sum_{i,j} |I(i,j)- \tilde I(i,j) |^2} {M.N} },$$
where $I$ is the groundtruth image, $ \tilde I$ is the restored one and $M$, $N$ are the image side lengths.
\section{Experiments}
\label{sec:experiments}
\subsection{Total variation based method}
Let $z_t(x,y)$ be the acquired image. The TV based method \cite{moisan} looks for a constant $k(y)$ to add at each column. So $$|| z_t(x,y)+k(y)||_{TV}$$ is as small as possible,
where $\label{def:TV}||I||_{TV}=\sum_{i,j}|(\nabla I)_{i,j}| $ and $(\nabla I)_{i,j}=
\begin{pmatrix}
I_{i+1,j} -I_{i,j}\\
I_{i+1,j}-I_{i,j}
\end{pmatrix}$.
So this amounts to the simple minimization of  $\sum_x|z_t(x,y+1)+\delta(y)-z_t(x,y)|$ for each column $y$.
 Then $k(y+1)=k(y)+\delta(y)$, where $k(0)=c$ chosen so that the resulting $I^{TV}_t$ and the input images $z_t$ have the same mean.\\

\subsection{Comparative experiments}
\begin{figure}[!h]
\begin{center}
\includegraphics[width=7cm]{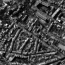}
\includegraphics[width=7cm]{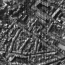}
\caption{\label{comp0}Image 1 : The groundtruth (left) the simulated FPN (right, RMSE=0.1932).}
\end{center}
\end{figure}

\begin{figure}[!h]
\begin{center}
\includegraphics[width=7cm]{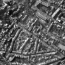}
\includegraphics[width=6.95cm]{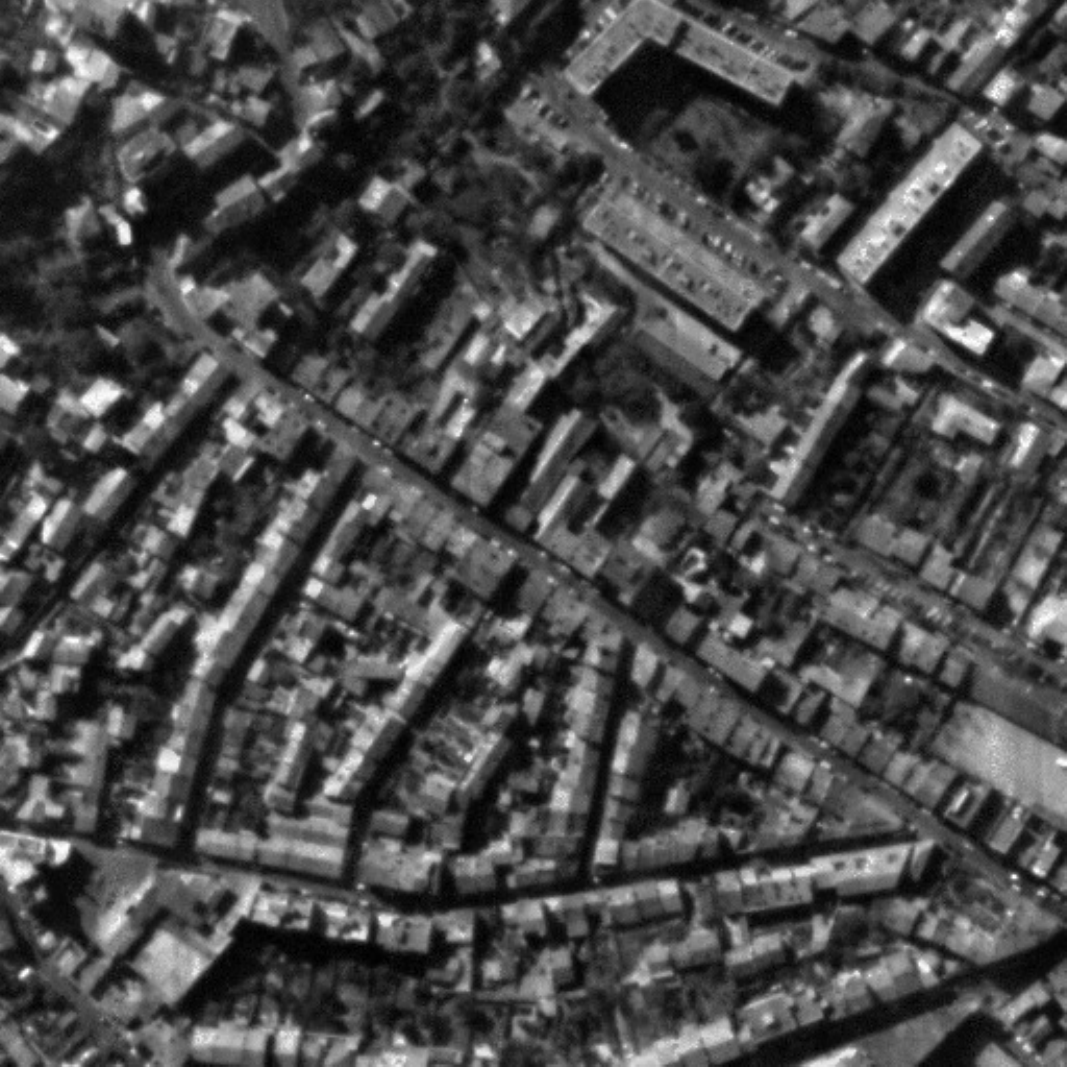}
\caption{\label{comp1}The TV based method (left, RMSE=0.1817), MIRE (right, RMSE=0.1715).}
\end{center}
\end{figure}

\begin{figure}[!h]
\begin{center}
\includegraphics[width=8cm]{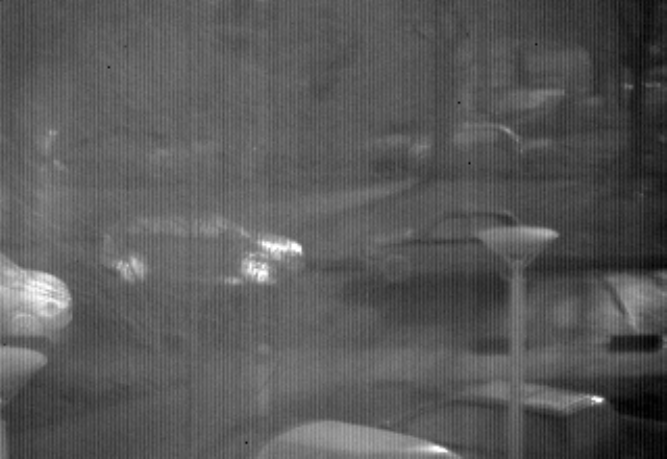}
\includegraphics[width=8cm]{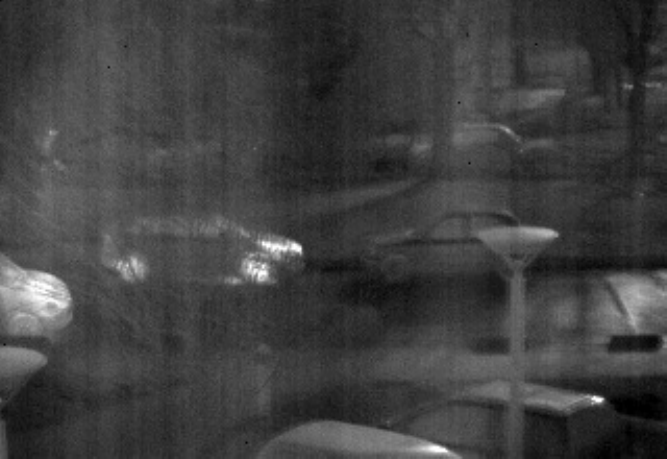}\\
\includegraphics[width=8cm]{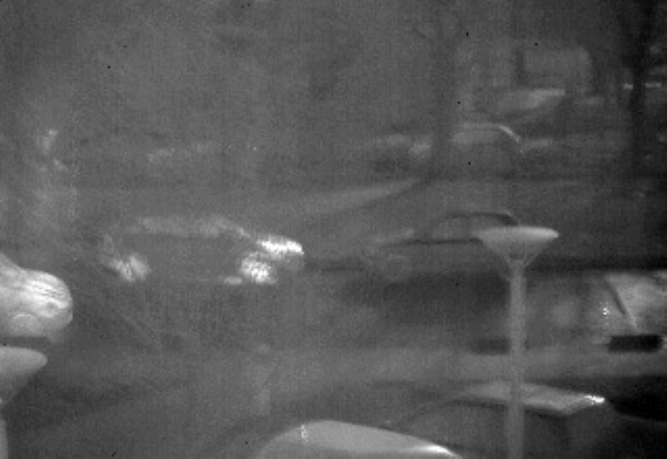}
\caption{\label{comp3}Top left : RAW (cooled camera), top right : TV based method, at the bottom : MIRE.}
\end{center}
\end{figure}

\begin{figure}[!h]
\begin{center}
\includegraphics[width=8cm]{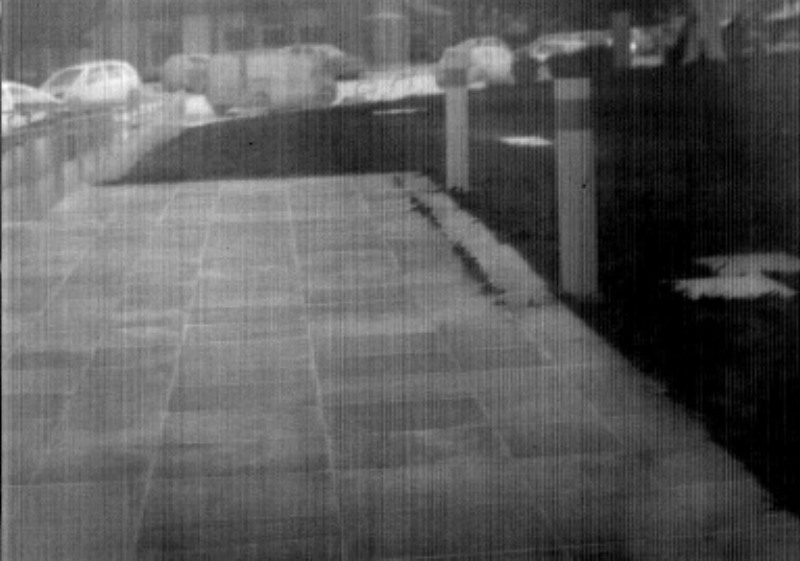}
\includegraphics[width=8cm]{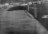}\\
\includegraphics[width=8cm]{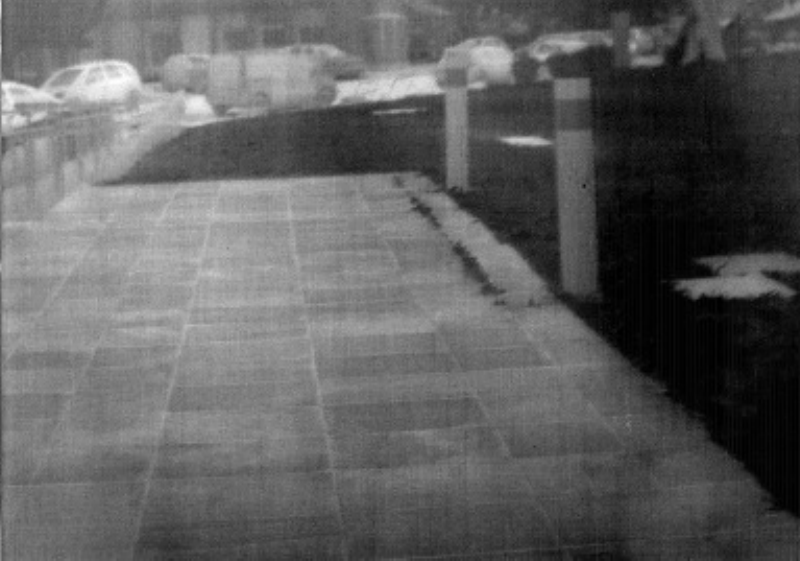}
\caption{\label{sim2}Top left : RAW (cooled camera), top right : TV based method, at the bottom : MIRE.}
\end{center}
\end{figure}

Simulations (Figs \ref{comp0}-\ref{comp1}) are made using a linear randomly generated model of NU. The comparative experiments of MIRE with Total Variation (TV) were processed using a \textit{Megawave \footnote{ \textit{M�gawave} is available at megawave.cmla.ens-cachan.fr/  }} (resthline  module \cite{moisan}). Results are quantified in term of RMSE and confirm the guess of visual improvement in quality.\\
Real experiments are shown using cooled (Fig \ref{comp3}) and uncooled (Fig \ref{sim2}) cameras. For comparison purpose images are shown with the same variance in every experiments. \\
MIRE always shows a significant improvement on TV and the final visual quality is overall very satisfactory. 

%\clearpage


\begin{figure}[!h]
\begin{center}
\includegraphics[width=4.5cm]{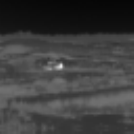}
\includegraphics[width=4.5cm]{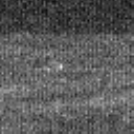}\\
\includegraphics[width=4.5cm]{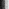}
\includegraphics[width=4.5cm]{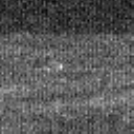}
\caption{\label{comp2}The only case where MIRE failed, on a textural 64x64 image.
Top-left the groundtruth, top right the observed (RMSE:0.7033).
Bottom left : the TV method (RMSE:1.2681), bottom right  : MIRE (RMSE:0.6833).}
\end{center}
\end{figure}

%\begin{figure}
%\begin{center}
%\begin{tabular}{|l|c|r|}
%  \hline
%  RMSE & TV delining & Us \\
%  \hline
%Image1 & 1.2 & 1.3 \\
% Image1 & 2.2 & 2.3 \\
%  \hline
%\end{tabular}
%\caption{Comparison in term of RMSE (simulated NU).}
%\end{center}
%\end{figure}
%\clearpage
\section{Discussion and conclusion}
\label{sec:conclusion}
\begin{figure}[!h]
\begin{center}
\includegraphics[width=8cm]{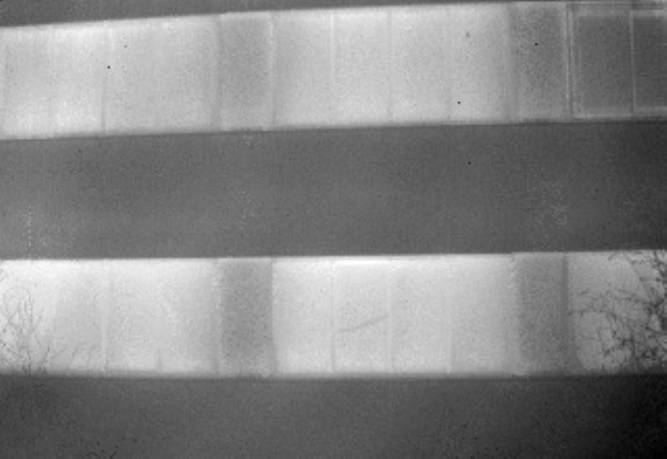}
\includegraphics[width=8cm]{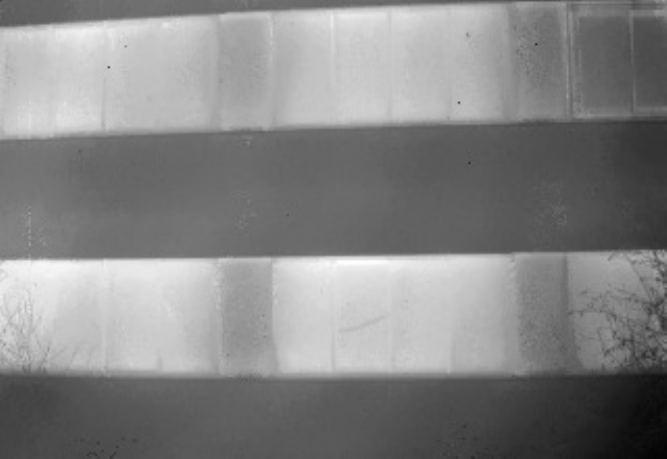}
\caption{\label{NLmeans}Before and after NL-means denoising, used with default parameters, available at www.ipol.im. Both are X2 zoomed.}
\end{center}
\end{figure}

In this paper a new way to correct for the uncooled IR  non-uniformity was proposed.  Evaluations using both simulated and real images --from both cooled and uncooled cameras-- show that the approach performs an efficient non-uniformity correction (in term of RMSE and visual image quality). Comparison was made with a total variation based method. This simple algorithm is well suited for a parallel implementation, since each column could be processed independently from the others. Furthermore since we process each image of the stream separately   "ghost artifacts" are not present and the velocity of the parameter drift insignificant.\\
Eventually the output seems to be more corrupted with gaussian temporal noise than with residues of unperfect correction of the non-uniformity. This enables the application of any standard image denoising algorithm, such as NL-Means or the wavelet thresholding, etc. See Fig \ref{NLmeans}.
The only failure case we met, shown in Fig \ref{comp2}, appeared with a small (64*64) simulated textured image. There were not enough bins in the histograms to equalize.
The results could still be enhanced by using a registration technique for badly corrupted images. This extension is envisaged in section \ref{sec:improvement}.

\clearpage
\section{Future work}
\label{sec:improvement}

\begin{figure}[!h]
\begin{center}
\includegraphics[width=5.5cm]{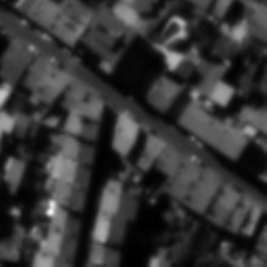}
\includegraphics[width=5.5cm]{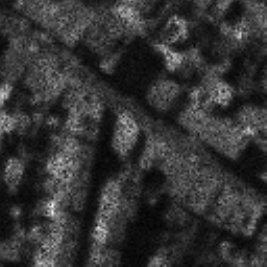}
\includegraphics[width=5.5cm]{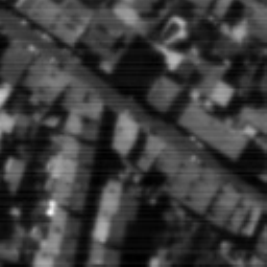}
\caption{\label{sim1}Using the motion, from left to right : the groundtruth, observed (one in the whole sequence),  restored.}
\end{center}
\end{figure}

Here is how MIRE could be combined with motion estimation:
\begin{enumerate}
\item First use MIRE
\item Apply a time registration
\item For each line parallel to the motion proceed as follows:
\label{propagation}\begin{itemize}
\item Choose a pixel as a reference.
\item Use motion to propagate this information in the direction of the motion.
\item Stop after the number of points is sufficient to the estimation of the non uniformity.
\end{itemize}
\end{enumerate}
At this step for each pixel sensor we have several points to estimate the transfer function. Hence we could perform any kind of interpolation to estimate a complete transfer function. Then we could compensate for linear as well as non linear uniformity since with $N$ images we will know up to $N$ points of the function. We don't need to perform a panorama to estimate the landscape like in \cite{hardie00}. It is another point of view on the problem, since these authors focus on estimating the landscape (difference of response between the perfect sensor and the real one) while the envisaged method is to obtain an estimation of the uniformity directly (more precisely an estimation of the difference of uniformity between an arbitrary sensor and the others).\\
 If we get an image with strong lining artifact in the motion direction, we have two possibilities : either to  use a new motion along another direction or using a single frame  algorithm like MIRE.\\
Fig \ref{sim1} presents some results on simulated images. We simulated a movie with a (pixelian) translational motion and a NU. The NU remained constant for the whole sequence and \textit{temporal} gaussian noise was added to each image. Then we applied step \ref{propagation} (assuming the motion is known).\\

\clearpage

\section{Appendix.}

\acknowledgments     %>>>> equivalent to \section*{ACKNOWLEDGMENTS}

We thank the D�l�gation G�n�rale pour l'Armement (DGA) for supporting this work.

\clearpage

%%%%%%%%%%%%%%%%%%%%%%%%%%%%%%%%%%%%%%%%%%%%%%%%%%%%%%%%%%%%%
%%%%% References %%%%%

\bibliography{biblio2}   %>>>> bibliography data in report.bib
\bibliographystyle{spiebib}   %>>>> makes bibtex use spiebib.bst
%\nocite{*}
\end{document}